\documentclass[letterpaper, 10 pt, conference]{ieeeconf}  

\IEEEoverridecommandlockouts                              

\usepackage{adjustbox}
\usepackage[ruled, vlined]{algorithm2e}
\usepackage{amsmath}
\usepackage{amssymb}
\usepackage{array}  
\usepackage{authblk}
\usepackage[accsupp]{axessibility}  
\usepackage{booktabs}
\usepackage{capt-of}
\makeatletter
\let\NAT@parse\undefined
\makeatother
\usepackage{cite}
\usepackage[font={small}]{caption}
\usepackage{colortbl}
\usepackage{epsfig}
\usepackage{inconsolata}  
\usepackage{graphicx}
\usepackage{listings}
\usepackage{multicol}
\usepackage{multirow}
\usepackage{paralist}
\usepackage{pifont}
\usepackage{times}
\usepackage{tabularx}
\usepackage{xcolor}
\usepackage{svg}

\definecolor{tabfirst}{rgb}{0, 0.25, 0}
\definecolor{tabthird}{rgb}{1, 0.85, 0.7}
\definecolor{tabsecond}{rgb}{1, 0.96, 0.7}

\definecolor{flodarkpurple}{rgb}{0.288,0.1196,0.7}

\definecolor{amber}{rgb}{1.0, 0.75, 0.0}

\usepackage[backref=page,breaklinks,colorlinks,bookmarks,citecolor=flodarkpurple]{hyperref}

\newcommand{\coolname}{\textit{3DIML}}

\newcommand{\authorhref}[3][flodarkpurple]{\href{#2}{\color{#1}{#3}}}

\title{\Large \bf
Efficient 3D Instance Mapping and Localization with Neural Fields 
}

\author{
\authorhref{https://gtangg12.github.io}{George Tang}, 
\authorhref{https://krrish94.github.io}{Krishna Murthy Jatavallabhula},
\authorhref{https://groups.csail.mit.edu/vision/torralbalab/}{Antonio Torralba}
\\
\href{https://www.csail.mit.edu/}{Massachusetts Institute of Technology}
}

\begin{document}


\makeatletter
\let\@oldmaketitle\@maketitle
\renewcommand{\@maketitle}{\@oldmaketitle
\centering
\includegraphics[width=0.975\linewidth]{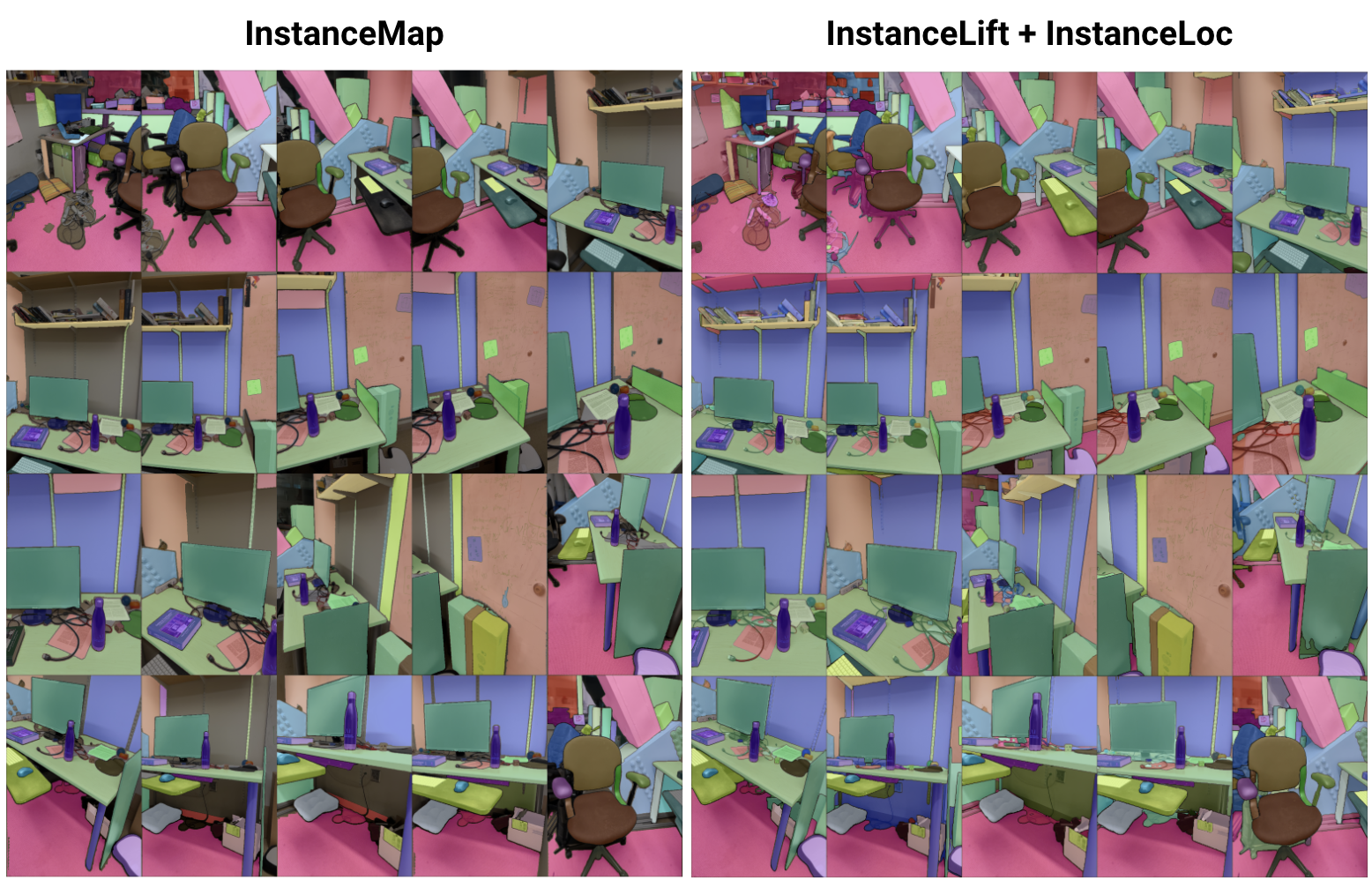}
\captionof{figure}{
\label{splash}
Our approach, \textbf{\coolname}, learns an implicit representation of a scene as a composition of object instances. It does so by lifting 2D view-inconsistent instance labels from off-the-shelf 2D segmentation models (such as the Segment Anything) into 3D view-consistent instance labels. The images above show results for the in-the-wild scan \texttt{postdoc office} generated using \coolname{}, composed of InstanceMap (left) and InstanceLift. InstanceLoc (right) is then used to refine the results. Each identified 3D label is shown in a different color. Notice how thin and partially occluded objects are accurately delineated across the sequence.
}
\label{fig:splash}
}
\makeatother

\maketitle
\thispagestyle{empty}
\pagestyle{empty}

\begin{abstract}

We tackle the problem of learning an implicit scene representation for 3D instance segmentation from a sequence of posed RGB images. Towards this, we introduce \coolname{}, a novel framework that efficiently learns a \emph{neural label field} which can render 3D instance segmentation masks from novel viewpoints. Opposed to prior art that optimizes a neural field in a self-supervised manner, requiring complicated training procedures and loss function design, \coolname{} leverages a two-phase process. The first phase, InstanceMap, takes as input 2D segmentation masks of the image sequence generated by a frontend instance segmentation model, and associates corresponding masks across images to 3D labels.
These almost 3D-consistent pseudolabel masks are then used in the second phase, InstanceLift, to supervise the training of a neural label field, which interpolates regions missed by InstanceMap and resolves ambiguities. 
Additionally, we introduce InstanceLoc, which enables near realtime localization of instance masks given a trained neural label field.
We evaluate \coolname{} on sequences from the Replica and ScanNet datasets and demonstrate its effectiveness under mild assumptions for the image sequences. We achieve a large practical speedup over existing implicit scene representation methods with comparable quality, showcasing its potential to facilitate faster and more effective 3D scene understanding.
\end{abstract}

\setcounter{figure}{1} 

\section{Introduction}
\label{sec:intro}

Intelligent agents require scene understanding at the object level to effectively carry out context-specific actions such as navigation and manipulation. While segmenting objects from images has seen remarkable progress with scalable models trained on internet-scale datasets~\cite{mask2former,sam}, extending such capabilites to the 3D setting remains challenging.

In this work, we tackle the problem of learning a 3D scene representation from posed 2D images that \textit{factorizes} the underlying scene into its set of constituent objects.
Existing approaches to tackle this problem have focused on training class-agnostic 3D segmentation models~\cite{mask3d,openmask3d}, requiring large amounts of annotated 3D data, and operating directly over explicit 3D scene representations (e.g., pointclouds).
An alternate class of approaches~\cite{panopli,contrastive} has instead proposed to directly \textit{lift} segmentation masks from off-the-shelf instance segmentation models into implicit 3D representations, such as neural radiance fields (NeRF)~\cite{nerf}, enabling them to render 3D-consistent instance masks from novel viewpoints.

However, the neural field-based approaches have remained notoriously difficult to optimize, with ~\cite{panopli} and ~\cite{contrastive} taking several hours to optimize for low-to-mid resolution images (e.g., $300 \times 640$).
In particular, Panoptic Lifting \cite{panopli} scales cubicly with the number of objects in the scene\, preventing it from being applied to scenes with hundreds of objects, while Contrastively Lifting \cite{contrastive} requires a complicated, multi-stage training procedure, hindering practicality for use in robotics applications.

To this end, we propose \coolname{}, an efficient technique to learn 3D-consistent instance segmentation from posed RGB images.
\coolname{} comprises two phases: InstanceMap and InstanceLift. Given view-inconsistent 2D instance masks extracted from the RGB sequence using a frontend instance segmentation model~\cite{sam}, InstanceMap produces a sequence of view-consistent instance masks.
To do so, we first associate masks across frames using keypoint matches between similar pairs of images.
We then use these potentially noisy associations to supervise a neural label field, InstanceLift, which exploits 3D structure to interpolate missing labels and resolve ambiguities. Unlike prior work, which requires multistage training and additional loss function engineering, we use a single rendering loss for instance label supervision, enabling the training process to converge significantly faster. The total runtime of \coolname{}, including InstanceMap, takes 10-20 minutes, as opposed to 3-6 hours for prior art.

In addition, we devise InstaLoc, a fast localization pipeline that takes in a novel view and localizes all instances segmented in that image (using a fast instance segmentation model~\cite{fastsam}) by sparsely querying the label field and fusing the label predictions with extracted image regions. 
Finally, \coolname{} is extremely modular, and we can easily swap components of our method for more performant ones as they become available.

To summarize, our contributions are:
\begin{itemize}
    \item An efficient neural field learning approach that factorizes a 3D scene into its constituent objects
    \item A fast instance localization algorithm that fuses sparse queries to the trained label field with performant image instance segmentation models to generate 3D-consistent instance segmentation masks
    \item A 12-25x reduction in the number of iterations needed to train the neural label field and practical runtime improvement of 14-24$\times$ over prior art, benchmarked on a single NVIDIA RTX 3090.
\end{itemize}

\section{Background}
\label{sec:related_work}

\begin{figure*}
    \centering
    \includegraphics[width=\linewidth]{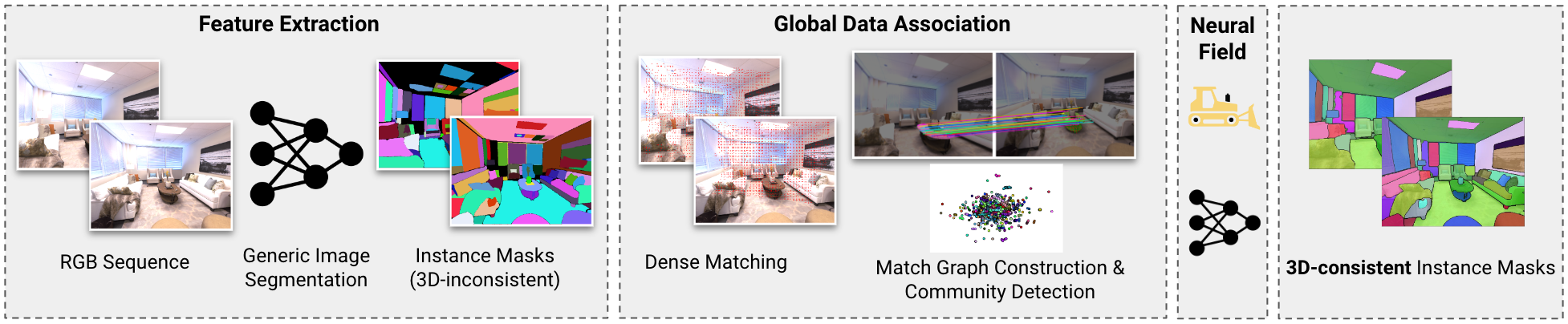}
    \caption{\textbf{Overview of \coolname{}}. A sequence of color images is segmented into object instances by an image segmentation backbone. The resulting masks produced are fed into InstanceMap, which produces instance masks consistent over all frames. These pseudo instance masks and their respective camera poses are used to supervise an instance label NeRF, which further improves consistency and resolves ambiguity present in the InstanceMap outputs. The \emph{feature extraction} and \emph{global data association} blocks together form InstanceMap.}
    \label{fig:instaML-pipeline}
\end{figure*}

\textbf{2D segmentation:}
The prevalence of vision transformer architecture and the increasing scale of image datasets have resulted in a series of state-of-the-art image segmentation models. Panoptic and Contrastive Lifting both lift panoptic segmentation masks produced by Mask2Former \cite{mask2former} to 3D by learning a neural field. 
Towards open-set segmentation, segment anything (SAM)~\cite{sam} achieves unprecedented performance by training on a billion masks over 11 million images. HQ-SAM~\cite{samhq} improves upon SAM for fine-grained masks. FastSAM~\cite{fastsam} distills SAM into a CNN architecture and achieves similar performance while being orders of magnitude faster. 
In this work, we use GroundedSAM~\cite{ram,groundingdino}, which combines SAM with Recognize Anything (RAM) \cite{ram} and GroundingDino \cite{groundingdino} to produce object-level, instead of part-level segmentation masks.

\textbf{Neural fields for 3D instance segmentation:}
NeRFs are implicit scene representations that can accurately encode complex geometry, semantics, and other modalities and resolve viewpoint inconsistent supervision~\cite{semanticnerf}. 
Panoptic lifting~\cite{panopli} constructs semantics and instances branches on an efficient variant of NeRF, TensoRF~\cite{tensorf}, utilizing a Hungarian matching loss function to assign learned instance masks to surrogate object IDs given reference view-inconsistent masks. This scales poorly with an increasing number of objects (owing to the cubic complexity of Hungarian matching).
Contrastive lifting~\cite{contrastive} addresses this by instead employing contrastive learning on scene features, with positive and negative relations determined by whether or not they project onto the same mask. 
In addition, contrastive lifting requires a slow-fast clustering-based loss for stable training, leading to faster performance than panoptic lifting but requires multiple stages of training, leading to slow convergence.
Concurrently to us, Instance-NeRF~\cite{instancenerf} directly learns a label field via supervision, but their instance labels are derived from the 3D object detector NeRF-RPN~\cite{nerfrpn}, which cannot detect small objects. On the contrary, our approach is based on lifting SAM, which allows us to segment objects at SAM's granularity.

\textbf{Structure from Motion:}
During mask association in InstanceMap, we take inspiration from scalable 3D reconstruction pipelines such as Hierarchical Localization (hLoc) \cite{hloc}. Specifically, we use visual descriptors produced by NetVLAD for matching image viewpoints as a first step, then perform keypoint extraction and matching using LoFTR \cite{loftr} as a preliminary for mask association on the matched image viewpoints.

\section{Method}
Given a sequence of $N$ posed RGB images, $(I_i, T_i)$ where $I$ denotes the image and $T$ pose, we first extract view-inconsistent instance masks $M_i$ using a generic instance segmentation model such as Mask2Former or SAM.  

\subsection{Mask Association}
We first generate pseudolabel masks with InstanceMap. Formally, define $\phi(M, r)$ to map a mask $M$ and region $r$ to a consistent label for the same 3D object across different masks and regions. We extend the popular hLoc \cite{hloc} framework for scalable 3D reconstruction to mask association as follows:

\textbf{Correspondence Generation:}
For each image $I_i$ in the input sequence, we compute NetVLAD \cite{hloc} feature descriptors. We then compute the cosine similarity among all pairs of images using these descriptors, defining a visual match as a similarity above a threshold of $\tau_{\text{global}}$. For each visual pair, we extract dense correspondences (i.e. set of keypoints in both images and a correspondence mapping from one set to the other) using LoFTR \cite{loftr} Since LoFTR computes keypoints independently per image pair, we aggregated keypoints from all pairs. Denote image $I_i$'s keypoints $k_i$ and matches between $I_i, I_j$ as $m_{ij}$. These correspondences are used in the next step to associate masks. 

Since NetVLAD and LoFTR don't have 3D information, \coolname{} only performs well if each image in the scan sequence contains enough context for these models. We observe empirically a good rule of thumb is to have at least one other recognizable landmark for frames containing near-identical objects.

\textbf{Mask Association Graph:}
Insofar, our approach produces instance masks and dense pixel correspondences among images that share a visual overlap. However, segmentation models such as SAM \cite{sam} suffer multiple issues: (a) segmentations of the same object need not be consistent across images, owing to viewpoint and appearance variations; and (b) owing to over-segmentation of objects, there isn't usually a one-one correspondence among masks.

To associate instance masks across the entire scene, we construct an \emph{mask association graph}, where the nodes are all instances detected across all images in the sequence. We then proceed to populate edges in this graph, corresponding to each pair of object instances that may be deemed a potential match. For each visual pair $(i, j)$, for all regions in $M_i$, we find a best-match region in $M_j$ if one exists. Let $r_u, r_v$ denote regions in $M_i, M_j$, $k_{i}(r_u)$, $k_{j}(r_v)$ to be the keypoints of $I_i, I_j$ contained in regions $r_u, r_v$, respectively, and $m_{ij}(r_u, r_v)$ be the matches between the regions. An edge with weight \textit{matching score} is added for $r_u, r_v$ if the matching score exceeds some threshold
$$\min \left( \frac{m_{ij}(r_u, r_v)}{|k_{i}(r_u)|}, \frac{m_{ij}(r_u, r_v)}{|k_{j}(r_v)|)} \right) > \tau_{\text{local}}$$

\textbf{Community Detection:}
Given this association graph, to remove noisy edges, we run community detection, specifically \texttt{igraph}'s implementation of the Leiden Algorithm with resolution parameter 0, over this graph and merge all masks in the same community (determined by threshold size $\tau_{\text{community}}$; the remaining masks are discarded). We also filter out all labels that do not feature in a minimum number of frames, as there are insufficient observations to accurately associate them across views. To account for potentially overlapping instances, a common occurrence due to perspective image formation, we compute the average mask area across all frames an instance exists in, which serves as a rasterization order for InstanceLift (instance masks with the largest area averaged over all frames are rasterized first). Example outputs are shown in Fig.~\ref{fig:splash}.

\subsection{Mask Refinement}
$\phi(M, r)$ is inherently noisy due to varying segmentation hierarchies for different instance masks due to differing viewpoints as well as design specifics. To address this, in InstanceLift we feed the pseudolabel masks to a label NeRF, which resolves some ambiguities. Still, NeRF cannot handle extreme cases of label ambiguity, to which we devise a fast post-processing method that determines and merges colliding labels based on random renders from the label NeRF. The few remaining, if any, ambiguities can be corrected via sparse human annotation.

\textbf{Neural Label Field}
Our label field represents the scene as a function $\Phi(x, d)$ that maps each point $\mathbf{x} \in \mathbb{R}^3$ and view-direction directions $\mathbf{d} \in \mathbb{S}^2$ to a corresponding density $\sigma(\mathbf{x})$ and other modalities $f(\textbf{x}, \textbf{d})$. These neural fields are queried via volumetric rendering by sampling per pixel rays across images, sampling points on rays, and estimating the volume rendering integral along those sampled points (see \cite{nerf} for more details). We base our NeRF on the off-the-shelf Nerfacto architecture \cite{nerfstudio}. We add a multiresolution hashgrid \cite{instantngp} for instance labels in addition to one for color and geometry as well as an instance prediction MLP to render instance segmentation logits, which we supervise with cross entropy using the pseudolabel masks from InstanceMap. As in \cite{panopli}, \cite{instancenerf}, we block gradient flow from the label grid to the geometry grid. We train our model to learn volume density, color, and instance labels simultaneously.

\textbf{Label Merging:}
Once the label NeRF is trained, we build a \textit{redundancy graph} where the nodes, $v$ denote labels and the number of edges, $|e|$ between nodes correlate with how likely the labels refer to the same object. Specifically, we randomly render $K$ views and compare the rendered instance mask with some mask generated on the fly by FastSAM. For each region of FastSAM, we add pairwise edges between all labels that occupy greater than $\tau_{area}$ fraction of the region. Post graph construction, we merge labels $a, b$ if
$$\frac{2|e_{ab}|}{\deg(v_a) + \deg(v_b)} > \tau_{merge}$$

Since we only need coarse information i.e. instance mask noise, we render images downsampled by a factor of 2.

\subsection{Fast Instance Localization and Rendering}
\begin{figure}
    \centering
    \includegraphics[width=\linewidth]{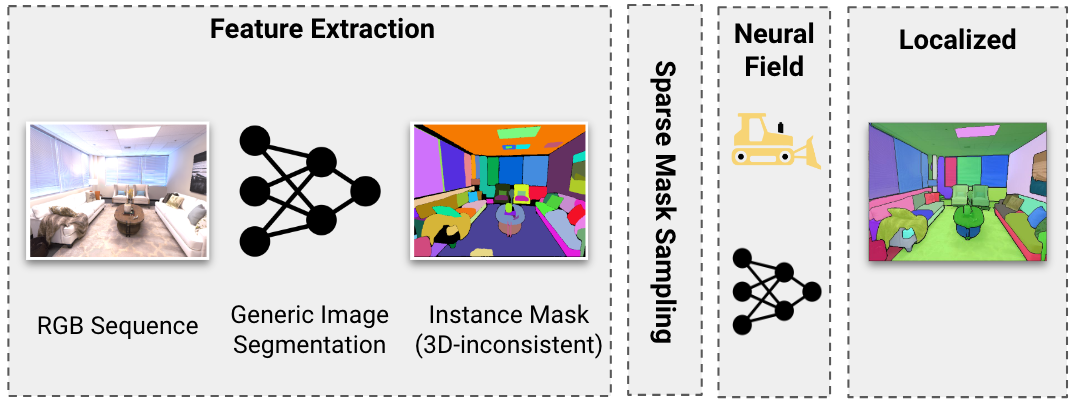}
    \caption{InstanceLoc enables 3D-consistent instance segmentation for novel views of the scene unobserved by the InstanceMap pipeline. We leverage off-the-shelf instance segmentation models to first produce 3D-inconsistent instance labels for a new input image. We then query the label field over a sparse set of points on the image and use this to localize each 2D instance mask i.e., assign a 3D-consistent label to each mask.
    }
    \label{fig:iML-localization}
\end{figure}

Training a label field enables us to predict 3D-consistent instance labels for novel viewpoints without rerunning \coolname{}. However, rendering every pixel is slow, and rendering from a novel viewpoint is often noisy. We propose a \emph{fast localization} approach that instead precomputes instance masks for the input image using an instance segmentation model (here FastSAM~\cite{fastsam}). Given this instance mask, for each instance region, we sample the corresponding pixelwise 3D object labels from the label NeRF and take the majority label. Another benefit is that the input instance masks can be constructed using prompts and edited before localization.

\section{Experiments}
\label{sec:results}
We benchmark our method against Panoptic and Contrastive Lifting using the same Mask2Former frontend. For fairness, we render semantics as in \cite{panopli}, \cite{contrastive} using the same multiresolution hashgrid for semantics and instances. For other experiments, we utilize GroundedSAM as our frontend and FastSAM for runtime performance critical tasks such as label merging and InstanceLoc.

\subsection{Datasets}
We evaluate our methods on a challenging subset of scans from Replica and ScanNet, which provide ground truth annotations. Since our methods are based on structure from motion techniques, we utilize the Replica-vMap~\cite{vmap} sequences, which is more indicative of real-world collected image sequences. For Replica and ScanNet, we avoid scans with various incompatibilities with our method (multi-room, low visibility, Nerfacto doesn't converge) as well as those containing many close-up views of identical objects that easily confuse NetVLAD and LoFTR.

\subsection{Metrics}
For lifting panoptic segmentation, we utilize Scene Level Panoptic Quality \cite{panopli}, defined as the Panoptic Quality for the concatenated sequence of images. For Grounded SAM, especially with instance masks for smaller objects, there is a divergence in alignment with ground truth annotations. As such, we report mIoU for predicted, reference masks that have IoU $> 0.5$ over all frames (TP in Scene Level Panoptic Quality) as well as the number of such matched masks and the total number of reference masks.

\subsection{Implementation Details}
For comparison with Panoptic and Contrastive Lifting, we follow their evaluation framework and map the Mask2Former outputs and ground truth semantic labels to 21 reduced ScanNet classes, considering only the "thing" subset of those classes. We train Panoptic and Contrastive Lifting using the code provided by \cite{contrastive}. We use the default parameterizations as their Replica experiment, which scales the number of iterations (128000) with the number of images. We build \coolname{} on top of the popular framework Nerfstudio \cite{nerfstudio} with \texttt{tinycudann} enabled \cite{instantngp}. Training using the pseudolabels produced by InstanceMap allows us to dramatically reduce the number of training iterations to 5000 iterations for Replica-vMap (both Mask2Former and GroundedSAM) and 10000 iterations for ScanNet. For our panoptic experiments, we use $\left( \tau_{\text{global}}, \tau_{\text{local}}, \tau_{\text{community}}, \tau_{\text{area}}, \tau_{\text{merge}}\right)=$ (0.25, 0.8, 2, 0.15, 0.75). For GroundedSAM, we use (0.25, 0.75, 3, 0.3, 0.6) for Replica-vMap and (0.4, 0.75, 3, 0.25, 0.75) for ScanNet.

\subsection{Results}
\begin{figure}
    \centering
    \includegraphics[width=\linewidth]{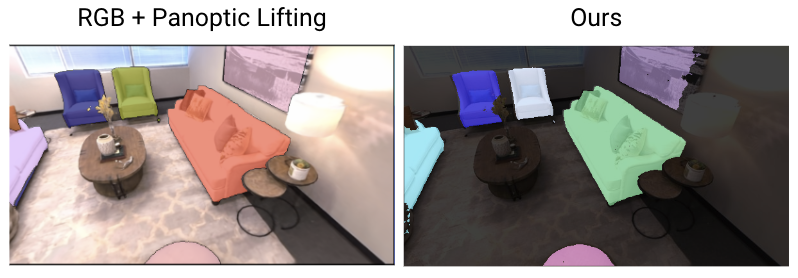}
    \caption{Comparison between Panoptic Lifting and \coolname{} for \texttt{room0} from Replica-vMap}
    \label{fig:pl}
\end{figure}

\textbf{Comparison with Panoptic and Contrastive Lifting:}
Table \ref{PQ} shows the Scene Level Panoptic Quality for \coolname{} and other methods on Replica-vMap sequences subsampled by 10 (200 frames). We observe \coolname{} approaches Panoptic Lifting in performance while achieving a much larger practical runtime (considering implementation) than Panoptic and Contrastive Lifting. Intuitively, this is achieved by efficiently relying on implicit scene representation methods only at critical junctions i.e. post InstanceMap, greatly reducing the number of training iterations of the neural field (25x less). Figure \ref{fig:pl} compares the instances identified by Panoptic Lifting to \coolname{}.

We benchmark all runtimes using a single RTX 3090 post-mask generation. Specifically, comparing their implementation to ours, Panoptic Lifting requires 5.7 hours of training over all scans, with a min and max of 3.6 and 6.6 hours, respectively, since its runtime depends on the number of objects. Contrastive Lifting takes around 3.5 hours on average while \coolname{} runs under 20 minutes (14.5 minutes on average) for all scans. Note several components of \coolname{} can be easily parallelized, such as dense descriptors extraction using LoFTR and label merging. The runtime of our method is dependent on the number of correspondences produced by LofTR, which doesn't change for different frontend segmentation models, and we observe similar runtimes for other experiments. 

\begin{table}[!h]
\centering
\adjustbox{max width=\linewidth}{%
\begin{tabular}{lcccccccc}
\toprule
\textbf{Scene} & office0 & office2 & office3 & office4 & room0 & room1 & \textbf{Average} \\
\midrule
Panoptic Lifting~\cite{panopli} &\cellcolor{tabsecond} 39.1 & \textbf{52.8} & \textbf{60.0} & \textbf{65.6} & \cellcolor{tabsecond} 63.7 & \textbf{94.8} & \textbf{62.6} \\
Contrastive Lift~\cite{contrastive} & 11.4 & 19.4 & 17.2 & 20.4 & 37.2 & \textbf{94.8} & 33.4 \\
InstanceMap (ours) &34.9 &26.1 &41.3 &27.3 &63.1 &52.5 & 40.8 \\
InstanceLift (ours) &\textbf{39.7} &\cellcolor{tabsecond} 28.1 & \cellcolor{tabsecond} 55.2 & \cellcolor{tabsecond} 46.4 & \textbf{68.8} & \cellcolor{tabsecond} 87.4 & \cellcolor{tabsecond} 54.2 \\
\bottomrule
\end{tabular}
} 
\caption{Quantitative comparison between pannoptic lifting~\cite{panopli}, contrastive lift~\cite{contrastive}, and our framework components, InstanceMap and InstanceLift. We measure the scene level panoptic quality metric (higher value indicates better performance). Our approach offers competitive performance while being far more efficient to train. Best performing numbers for each scene are in bold, while the second-best numbers are shaded \colorbox{tabsecond}{yellow}.}
\label{PQ}
\end{table}

\begin{table}[!h]
\centering
\adjustbox{max width=\linewidth}{%
\begin{tabular}{ll}
\toprule
Method & Time \\
\midrule
Panoptic lifting~\cite{panopli} & \ 5.7 hours \\
Contrastive lift~\cite{contrastive} & \ 3.6 hours \\
\midrule
InstanceMap (LoFTR) & \ 6.2 minutes \\
InstanceMap & \ 2.1 minutes \\
InstanceLift (train) & \ 5.6 minutes \\
InstanceLift (refine) & 33.2 seconds \\
\coolname{} (total) & \textbf{14.5 minutes} \\
\bottomrule
\end{tabular}
} 
\caption{Runtime in minutes benchmarked on a single RTX 3090 of Panoptic Lifting, Contrastive Lifting, and \coolname{}}
\label{PQ}
\end{table}

\begin{figure}[!thb]
    \centering
    \includegraphics[width=\linewidth]{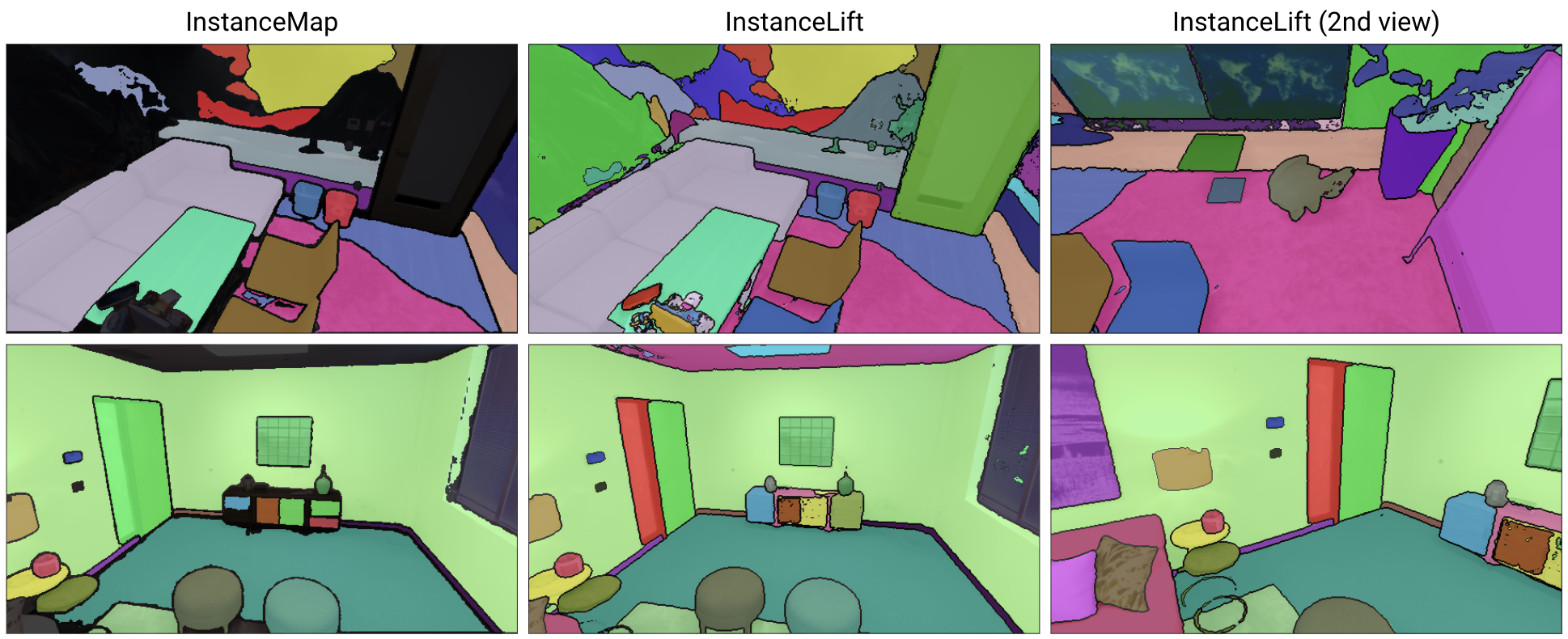}
    \caption{
    InstanceLift is able to fill in labels missed by InstanceMap as well as correct ambiguities. Here we show comparisons between them for \texttt{office0} and \texttt{room0} from Replica-vMap.
    }
    \label{fig:InstanceLift}
\end{figure}

\begin{figure}[!thb]
    \centering
    \includegraphics[width=\linewidth]{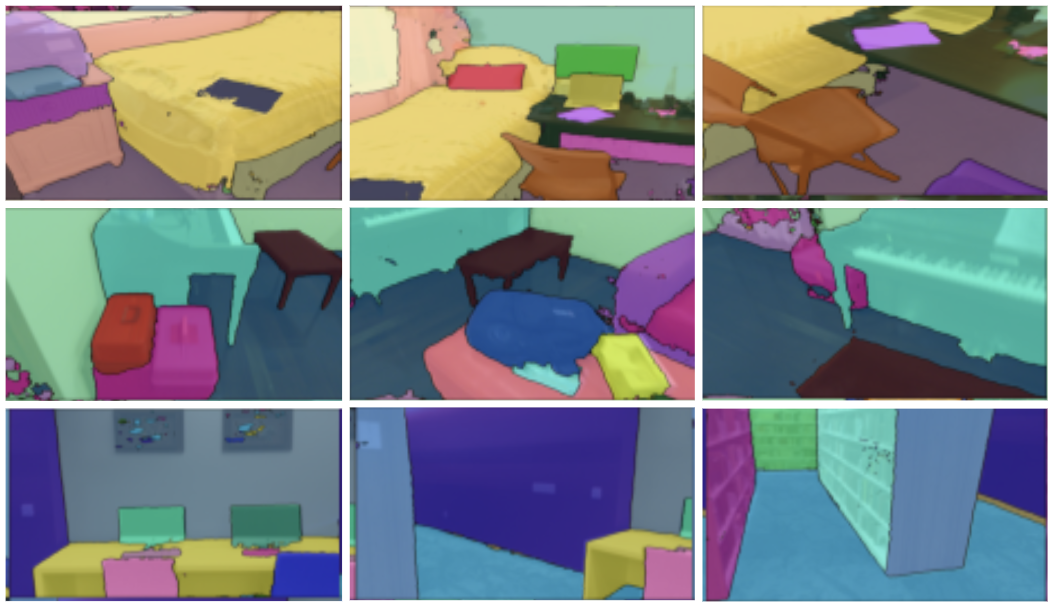}
    \caption{
    Some results for scans \texttt{0144\_01}, \texttt{0050\_02}, and \texttt{0300\_01} from ScanNet~\cite{scannet} (one scene per row, top to bottom), showcasing how \coolname{} accurately and consistently delineates instances in 3D.
    }
    \label{fig:scannet}
\end{figure}

\begin{figure*}
    \centering
    \includegraphics[width=\linewidth]{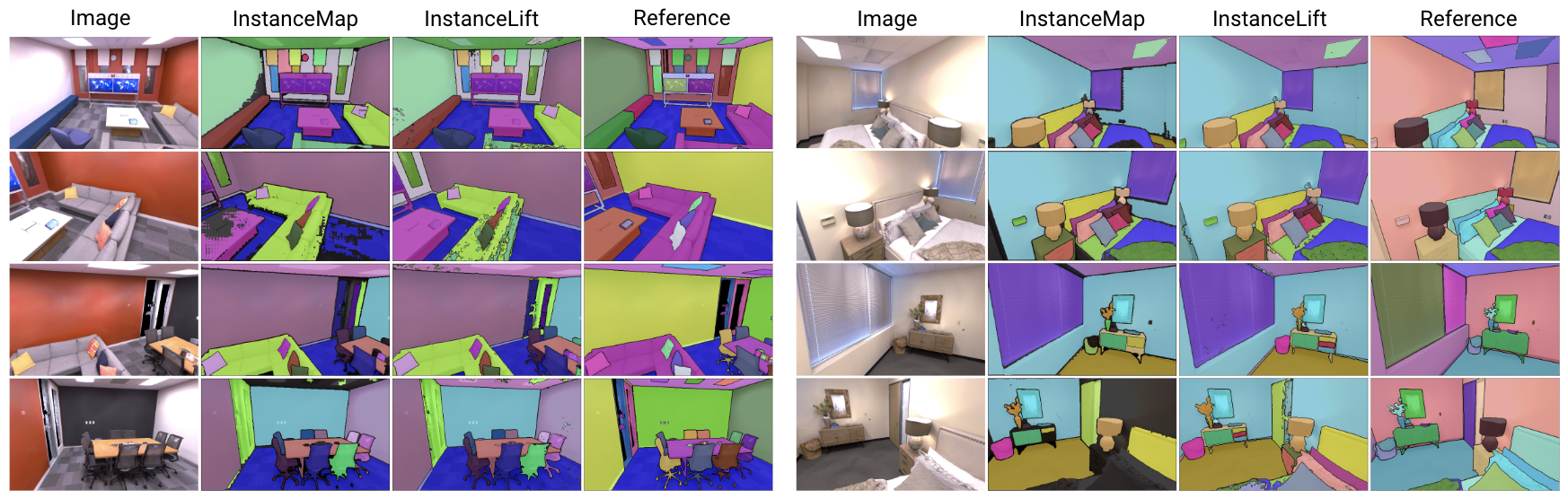}
    \caption{
    \textbf{Qualitative results} on \texttt{office3} and \texttt{room1} from the Replica-vMap split~\cite{vmap}. Both InstanceMap and InstanceLift are able to maintain quality and consistency over the image sequence despite duplicate objects due to sufficient image context overlap across the sequence.
    }
    \label{fig:replica}
\end{figure*}

\textbf{Grounded SAM}
Table \ref{replica mIoU} shows our results for lifting GroundedSAM masks for Replica-vMap. From Figure \ref{fig:InstanceLift} we see that InstanceLift is effective at interpolating labels missed by InstanceMap and resolving ambiguities produced by GroundedSAM \footnote{GroundedSAM produces lower quality frontend masks than SAM due to prompting using bounding boxes instead of a point grid.}. Figure \ref{fig:replica} shows that InstanceMap and \coolname{} are robust to large viewpoint changes and as well as duplicate objects, assuming nice scans, that is enough context for NetVLAD and LoFTR to somewhat distinguish between them. Table \ref{scannet mIoU} and Fig.~\ref{fig:scannet} illustrate our performance on ScanNet~\cite{scannet}.

\begin{table*}[!h]
\centering
\adjustbox{max width=\linewidth}{%
\begin{tabular}{lrrrrrrrr}\toprule
Scene &office0 &office2 &office3 &office4 &room0 &room1 &Average \\
\midrule
InstanceMap &0.740 / 14 &0.753 / 27 &0.769 / 30 &0.758 / 22 &0.825 / 34 &0.729 / 25 &0.762 \\
InstanceLift &0.785 / 18 &0.780 / 30 &0.799 / 36 &0.794 / 26 &0.851 / 38 &0.769 / 25 &0.796 \\
\bottomrule
\end{tabular}
} 
\caption{
Quantitative (mIoU, TP) results for GroundedSAM frontend on Replica-vMap. The average number of reference instances for all Replica scenes we evaluated on is $67$.
}
\label{replica mIoU}
\end{table*}

\begin{table*}[!h]
\centering
\adjustbox{max width=\linewidth}{%
\begin{tabular}{lrrrrrrrrr}\toprule
Scene &0050\_02 &0144\_01 &0221\_01 &0300\_01 &0389\_00 &0494\_00 &0693\_00 &Average \\
\midrule
InstanceMap &0.675 / 12 &0.646 / 14 &0.667 / 12 &0.726 / 13 &0.704 / 13 &0.716 / 6 &0.700 / 14 &0.691 \\
InstanceLift &0.657 / 14 &0.664 / 14 &0.621 / 10 &0.754 / 14 &0.745 / 12 &0.712 / 8 &0.782 / 16 &0.705 \\
\bottomrule
\end{tabular}
} 
\caption{
Quantitative (mIoU, TP) results for GroundedSAM frontend on ScanNet. The average number of reference instances for all ScanNet scenes we evaluated on is $32$.
}
\label{scannet mIoU}
\end{table*}

\textbf{Novel View Rendering and InstanceLoc}
Table \ref{novel} shows the performance if \coolname{} on the second track provided in Replica-vMap. We observe that InstanceLift is effective at rendering new views, and therefore InstanceLoc performs well. For Replica-vMap and FastSAM, InstanceLoc takes on average 0.16s per localized image (6.2 frames per second). In addition, InstanceLoc can be applied as a post-processing step to the renders of the input sequence as a denoising operation. 

\begin{table*}[!h]
\centering
\adjustbox{max width=\linewidth}{%
\begin{tabular}{lrrrrrrr}\toprule
Scene &office0 &office2 &office3 &office4 &room0 &room1 &Average \\
\midrule
InstanceLift &0.822 / 16 &0.787 / 29 &0.762 / 35 &0.769 / 30 &0.849 / 39 &0.768 / 28 & 0.793 \\
InstanceLoc &0.845 / 15 &0.842 / 30 &0.799 / 37 &0.812 / 30 &0.877 / 41 &0.800 / 28 & 0.829 \\
\bottomrule
\end{tabular}
} 
\caption{
Quantitative (mIoU, TP) results for InstanceLift and InstanceLoc on novel views over the Replica-vMap split~\cite{vmap}. 
}
\label{novel}
\end{table*}

\begin{figure}
    \centering
    \includegraphics[width=\linewidth]{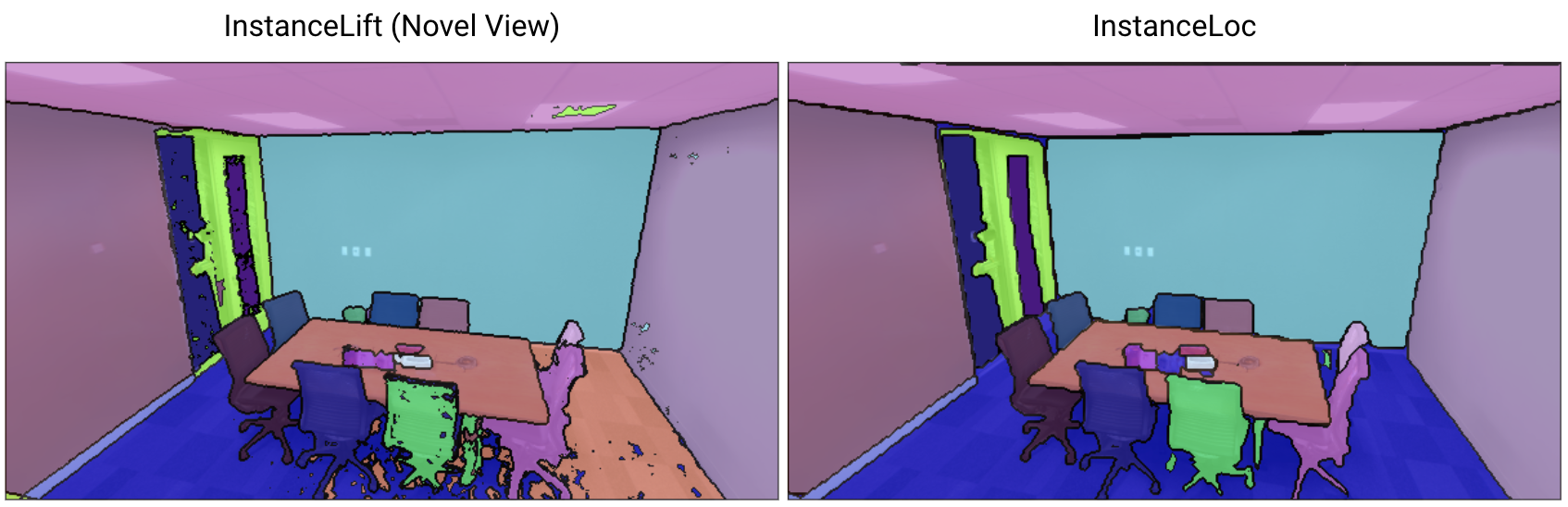}
    \caption{
    InstanceLoc is able to correct for noise rendered by InstanceLift.
    }
    \label{fig:InstanceLoc}
\end{figure}

\subsection{Limitations and Future Work}
In extreme viewpoint changes, our method sometimes produced discontinuous 3D instance labels. For example, on the worst performing scene, \texttt{office2}, we see that since the scan images only the front of a chair facing the back of the room and the back of a chair facing the front of the room for many frames, InstanceMap is not able to conclude these labels refer to the same object, and InstanceLift was unable to fix it, as NeRF's correction performance rapidly degrades with increasing label inconsistency \cite{semanticnerf}. However, there are very few of these left per scene post \coolname{}, and they can be easily fixed via sparse human annotation.

\begin{figure}
    \centering
    \includegraphics[width=\linewidth]{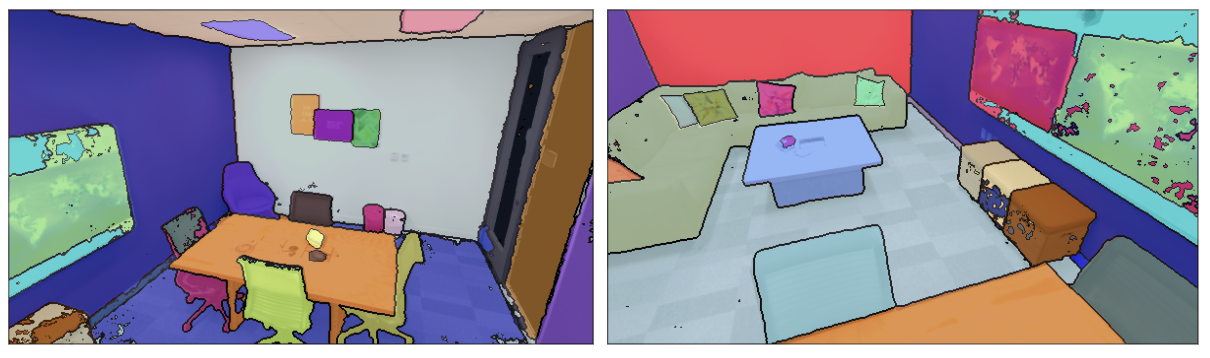}
    \caption{
    Our methods do not perform well in cases where the scan sequence contains only images of the different sides of an object (chair) or surface (floor) from differing directions without any smooth transitions in between, which occurs for \texttt{office2} from Replica (vMap split~\cite{vmap}).
    }
    \label{fig:failure}
\end{figure}

\section{Conclusion}

In this paper, we present \coolname{}, which addresses the problem of 3D instance segmentation in a class-agnostic and computationally efficient manner. By employing a novel approach that utilizes InstanceMap and InstanceLift for generating and refining view-consistent pseudo instance masks from a sequence of posed RGB images, we circumvent the complexities associated with previous methods that only optimize a neural field. Furthermore, the introduction of InstanceLoc allows rapid localization of instances in unseen views by combining fast segmentation models and a refined neural label field. Our evaluations across Replica and ScanNet and different frontend segmentation models showcase \coolname{}'s speed and effectiveness. It offers a promising avenue for real-world applications requiring efficient and accurate scene analysis.



\bibliographystyle{IEEEtran}
\bibliography{IEEEtranBST/IEEEabrv, root}

\begin{thebibliography}{10}
\providecommand{\url}[1]{#1}
\csname url@samestyle\endcsname
\providecommand{\newblock}{\relax}
\providecommand{\bibinfo}[2]{#2}
\providecommand{\BIBentrySTDinterwordspacing}{\spaceskip=0pt\relax}
\providecommand{\BIBentryALTinterwordstretchfactor}{4}
\providecommand{\BIBentryALTinterwordspacing}{\spaceskip=\fontdimen2\font plus
\BIBentryALTinterwordstretchfactor\fontdimen3\font minus
  \fontdimen4\font\relax}
\providecommand{\BIBforeignlanguage}[2]{{%
\expandafter\ifx\csname l@#1\endcsname\relax
\typeout{** WARNING: IEEEtran.bst: No hyphenation pattern has been}%
\typeout{** loaded for the language `#1'. Using the pattern for}%
\typeout{** the default language instead.}%
\else
\language=\csname l@#1\endcsname
\fi
#2}}
\providecommand{\BIBdecl}{\relax}
\BIBdecl

\bibitem{mask2former}
\BIBentryALTinterwordspacing
B.~Cheng, I.~Misra, A.~G. Schwing, A.~Kirillov, and R.~Girdhar,
  ``Masked-attention mask transformer for universal image segmentation,''
  \emph{CoRR}, vol. abs/2112.01527, 2021. [Online]. Available:
  \url{https://arxiv.org/abs/2112.01527}
\BIBentrySTDinterwordspacing

\bibitem{sam}
A.~Kirillov, E.~Mintun, N.~Ravi, H.~Mao, C.~Rolland, L.~Gustafson, T.~Xiao,
  S.~Whitehead, A.~C. Berg, W.-Y. Lo \emph{et~al.}, ``Segment anything,''
  \emph{arXiv preprint arXiv:2304.02643}, 2023.

\bibitem{mask3d}
J.~Schult, F.~Engelmann, A.~Hermans, O.~Litany, S.~Tang, and B.~Leibe,
  ``Mask3d: Mask transformer for 3d semantic instance segmentation,'' in
  \emph{2023 IEEE International Conference on Robotics and Automation
  (ICRA)}.\hskip 1em plus 0.5em minus 0.4em\relax IEEE, 2023, pp. 8216--8223.

\bibitem{openmask3d}
A.~Takmaz, E.~Fedele, R.~W. Sumner, M.~Pollefeys, F.~Tombari, and F.~Engelmann,
  ``Openmask3d: Open-vocabulary 3d instance segmentation,'' \emph{arXiv
  preprint arXiv:2306.13631}, 2023.

\bibitem{panopli}
Y.~Siddiqui, L.~Porzi, S.~R. Bul{\`o}, N.~M{\"u}ller, M.~Nie{\ss}ner, A.~Dai,
  and P.~Kontschieder, ``Panoptic lifting for 3d scene understanding with
  neural fields,'' in \emph{Proceedings of the IEEE/CVF Conference on Computer
  Vision and Pattern Recognition}, 2023, pp. 9043--9052.

\bibitem{contrastive}
Y.~Bhalgat, I.~Laina, J.~F. Henriques, A.~Zisserman, and A.~Vedaldi,
  ``Contrastive lift: 3d object instance segmentation by slow-fast contrastive
  fusion,'' 2023.

\bibitem{nerf}
B.~Mildenhall, P.~P. Srinivasan, M.~Tancik, J.~T. Barron, R.~Ramamoorthi, and
  R.~Ng, ``Nerf: Representing scenes as neural radiance fields for view
  synthesis,'' \emph{Communications of the ACM}, vol.~65, no.~1, pp. 99--106,
  2021.

\bibitem{fastsam}
X.~Zhao, W.~Ding, Y.~An, Y.~Du, T.~Yu, M.~Li, M.~Tang, and J.~Wang, ``Fast
  segment anything,'' \emph{arXiv preprint arXiv:2306.12156}, 2023.

\bibitem{samhq}
L.~Ke, M.~Ye, M.~Danelljan, Y.~Liu, Y.-W. Tai, C.-K. Tang, and F.~Yu, ``Segment
  anything in high quality,'' \emph{arXiv preprint arXiv:2306.01567}, 2023.

\bibitem{ram}
Y.~Zhang, X.~Huang, J.~Ma, Z.~Li, Z.~Luo, Y.~Xie, Y.~Qin, T.~Luo, Y.~Li,
  S.~Liu, Y.~Guo, and L.~Zhang, ``Recognize anything: A strong image tagging
  model,'' 2023.

\bibitem{groundingdino}
S.~Liu, Z.~Zeng, T.~Ren, F.~Li, H.~Zhang, J.~Yang, C.~Li, J.~Yang, H.~Su,
  J.~Zhu, and L.~Zhang, ``Grounding dino: Marrying dino with grounded
  pre-training for open-set object detection,'' 2023.

\bibitem{semanticnerf}
S.~Zhi, T.~Laidlow, S.~Leutenegger, and A.~J. Davison, ``In-place scene
  labelling and understanding with implicit scene representation,'' in
  \emph{Proceedings of the IEEE/CVF International Conference on Computer
  Vision}, 2021, pp. 15\,838--15\,847.

\bibitem{tensorf}
A.~Chen, Z.~Xu, A.~Geiger, J.~Yu, and H.~Su, ``Tensorf: Tensorial radiance
  fields,'' in \emph{European Conference on Computer Vision (ECCV)}, 2022.

\bibitem{instancenerf}
B.~Hu, J.~Huang, Y.~Liu, Y.-W. Tai, and C.-K. Tang, ``Instance neural radiance
  field,'' \emph{arXiv preprint arXiv:2304.04395}, 2023.

\bibitem{nerfrpn}
------, ``Nerf-rpn: A general framework for object detection in nerfs,'' in
  \emph{Proceedings of the IEEE/CVF Conference on Computer Vision and Pattern
  Recognition}, 2023, pp. 23\,528--23\,538.

\bibitem{hloc}
P.-E. Sarlin, C.~Cadena, R.~Siegwart, and M.~Dymczyk, ``From coarse to fine:
  Robust hierarchical localization at large scale,'' in \emph{Proceedings of
  the IEEE/CVF Conference on Computer Vision and Pattern Recognition}, 2019,
  pp. 12\,716--12\,725.

\bibitem{loftr}
J.~Sun, Z.~Shen, Y.~Wang, H.~Bao, and X.~Zhou, ``Loftr: Detector-free local
  feature matching with transformers,'' in \emph{Proceedings of the IEEE/CVF
  conference on computer vision and pattern recognition}, 2021, pp. 8922--8931.

\bibitem{nerfstudio}
M.~Tancik, E.~Weber, E.~Ng, R.~Li, B.~Yi, T.~Wang, A.~Kristoffersen, J.~Austin,
  K.~Salahi, A.~Ahuja \emph{et~al.}, ``Nerfstudio: A modular framework for
  neural radiance field development,'' in \emph{ACM SIGGRAPH 2023 Conference
  Proceedings}, 2023, pp. 1--12.

\bibitem{instantngp}
\BIBentryALTinterwordspacing
T.~Müller, A.~Evans, C.~Schied, and A.~Keller, ``Instant neural graphics
  primitives with a multiresolution hash encoding,'' \emph{ACM Transactions on
  Graphics}, vol.~41, no.~4, p. 1–15, Jul. 2022. [Online]. Available:
  \url{http://dx.doi.org/10.1145/3528223.3530127}
\BIBentrySTDinterwordspacing

\bibitem{vmap}
X.~Kong, S.~Liu, M.~Taher, and A.~J. Davison, ``vmap: Vectorised object mapping
  for neural field slam,'' in \emph{Proceedings of the IEEE/CVF Conference on
  Computer Vision and Pattern Recognition}, 2023, pp. 952--961.

\bibitem{scannet}
A.~Dai, A.~X. Chang, M.~Savva, M.~Halber, T.~Funkhouser, and M.~Nie{\ss}ner,
  ``Scannet: Richly-annotated 3d reconstructions of indoor scenes,'' in
  \emph{Proc. Computer Vision and Pattern Recognition (CVPR), IEEE}, 2017.

\end{thebibliography}

\end{document}